\documentclass{article}

\usepackage[nonatbib,preprint]{neurips_2021}

\usepackage[utf8]{inputenc} 
\usepackage[T1]{fontenc}    
\usepackage{xr-hyper}
\usepackage{hyperref}       
\usepackage{url}            
\usepackage{booktabs}       
\usepackage{amsfonts}       
\usepackage{nicefrac}       
\usepackage{microtype}      
\usepackage{xcolor}         

\usepackage{times}
\usepackage{epsfig}
\usepackage{graphicx}
\usepackage{amsmath}
\usepackage{amssymb}
\usepackage{subfig}
\usepackage{comment}

\usepackage{float}
\usepackage{booktabs} 
\usepackage{multirow}
\usepackage{multicol}

\usepackage{xspace}

\makeatletter
\DeclareRobustCommand\onedot{\futurelet\@let@token\@onedot}
\def\@onedot{\ifx\@let@token.\else.\null\fi\xspace}

\def\eg{\emph{e.g}\onedot}

\def\wrt{w.r.t\onedot} 

\makeatother

\newcommand\given{\medspace|\medspace}
\newcommand\argmax{\operatornamewithlimits{argmax}}

\makeatletter
\newcommand*{\addFileDependency}[1]{
  \typeout{(#1)}
  \@addtofilelist{#1}
  \IfFileExists{#1}{}{\typeout{No file #1.}}
}
\makeatother

\newcommand*{\myexternaldocument}[1]{%
    \externaldocument{#1}%
    \addFileDependency{#1.tex}%
    \addFileDependency{#1.aux}%
}
\myexternaldocument{probabilistic_attention_appendix}

\title{Probabilistic Attention for Interactive Segmentation}

\author{%
    Prasad Gabbur \\
    Apple \\
    \texttt{pgabbur@apple.com}
    \And
    Manjot Bilkhu \\
    Apple \\
    \texttt{mbilkhu@apple.com}
    \And
    Javier Movellan \\
    Apple \\
    \texttt{movellan@apple.com}
}

\begin{document}

\maketitle

\begin{abstract}
We provide a probabilistic  
interpretation of attention and show that the standard dot-product attention in transformers 
is a special case of Maximum A Posteriori (MAP) inference. The proposed approach suggests the use of Expectation Maximization algorithms for on-line adaptation of key and value  model parameters. This approach is  useful for cases in which external agents, e.g.,  annotators, provide inference-time information about the correct values of some tokens, e.g, the semantic category of some pixels,  and
we need for this new information to propagate to other tokens in a principled manner. We illustrate the approach on an interactive semantic segmentation 
task in which 
annotators and models collaborate on-line to improve 
	annotation efficiency. Using standard benchmarks, we observe that key adaptation boosts model performance ($\sim10\%$ mIoU) in the low feedback regime and value propagation improves model responsiveness in the high feedback regime. A PyTorch layer implementation of our probabilistic attention model will be made publicly available here: \url{https://github.com/apple/ml-probabilistic-attention}. 
\end{abstract}

\section{Introduction}
Attention was first introduced as a computational primitive for natural language processing \cite{Vaswani_2017_NIPS} and has since been widely adopted \cite{Devlin_2019_ACL, Yang_2019_NIPS, Dai_2019_ACL, Dehghani_2019_ICLR} as a replacement for recurrent primitives such as LSTMs \cite{Hochreiter_1997_NC}. More recently it has been making inroads into computer vision \cite{ramachandran2019standalone, Zhao_2020_CVPR, Wang_2018_CVPR, Yin_2020_ECCV, Bello_2019_ICCV, Dosovitskiy_2020_ArXiv, Srinivas_2021_arxiv} as a replacement for the long accepted convolution as the main computational primitive. Self-attention based architectures have demonstrated state-of-the-art results in fundamental vision problems including image classification \cite{Bello_2019_ICCV, ramachandran2019standalone, Zhao_2020_CVPR, Dosovitskiy_2020_ArXiv, Srinivas_2021_arxiv}, object detection \cite{Carion_2020_ECCV, Zhu_2021_ICLR, Wang_2018_CVPR}, image and video semantic segmentation \cite{Wang_2020_ECCV, Huang_2019_ICCV, Wang_2018_CVPR, Oh_2020_PAMI} and tracking \cite{Yu_2020_CVPR} to state a few.  

There are a few different perspectives on the reasons for success of self-attention in computer vision and its superiority over convolution. This includes a view that the self-attention mechanism allows modeling spatially varying dynamic convolution filters \cite{Li_2021_CVPR} and at the same time enabling parameter independent scaling of receptive fields \cite{Vaswani_2021_ArXiv}. Another includes their ability to capture global context through long range interactions especially when full attention is feasible \cite{Srinivas_2021_arxiv} at reduced spatial resolution maps or using an approximation of full attention with axial \cite{Wang_2020_ECCV} or criss-cross attention \cite{Huang_2019_ICCV}. A recent work \cite{Ramsauer_2021_ICLR} introduces modern Hopfield networks with continuous states where the update mechanism is shown to be equivalent to the update mechanism of standard dot-product attention \cite{Vaswani_2017_NIPS}. They show that such a network has the capacity to store exponentially many patterns and retrieve them with high fidelity.
In this work, we provide a novel interpretation of attention as a probabilistic generative model for queries and values.  Specifically we hypothesize the existence of a bank of probabilistic memory units, each of which maintains a joint probability distribution over queries and values parameterized through keys. A query/value pair is generated by first sampling a unit (from a prior over units) followed by sampling the pair from the unit specific joint distribution. This is equivalent to generating the queries and values through a probabilistic mixture model over the units. A particular form for unit joint likelihoods expressed as Gaussians for both the query and value marginals, assuming their independence conditioned on a unit, turns out to be equivalent to traditional dot product attention under a few constraints. 
As shown in Section~\ref{sec:map_value_inference}, maximum likelihood (ML) inference for the corresponding value given a query is equivalent to standard dot-product attention.

Our probabilistic interpretation provides a systematic framework for online update of mixture model parameters based on a set of observed queries. It also allows propagation of  correct values provided by an external agent for some of the units to all other units. Using 
Bayesian inference in the constrained case, 
we derive update rules for \textit{online unsupervised adaptation} (Section~\ref{sec:unsupervised_model_adaptation}) of query/key likelihood parameters based on a set of observed queries. We also derive update equations for \textit{online value propagation} (Section~\ref{sec:online_value_propagation}) across units based on fixed externally specified values for a subset of units. The latter is specifically useful for interactive segmentation where a correction provided by an annotator has to be propagated globally to make the process more efficient. We use probabilistic attention in place of standard attention in deep architectures for interactive segmentation both within the backbone and at the network head as a classifier. Specifically we use probabilistic attention updates in the BoTNet50 \cite{Srinivas_2021_arxiv} architecture and show that adapting keys to incoming queries leads to better model 
performance in the low annotator feedback regime. Using value propagation within a probabilistic attention layer at the head of the segmentation network leads to a more responsive model through effective feedback propagation in the high feedback regime. We also use both key adaptation and value propagation together and demonstrate the complementary effects of the two in both the low and high annotator feedback regimes. 

\section{Related Work}
\subsection{Attention}
Natural language processing has seen the rise \cite{Bahdanau_2014_ArXiv, Vaswani_2017_NIPS} and widespread adoption of attention in recent years \cite{Devlin_2019_ACL, Yang_2019_NIPS, Dai_2019_ACL, Dehghani_2019_ICLR, Wu_2019_ICLR}. One of the first works on visual attention was on learning to attend to image regions for caption generation \cite{Xu_2017_BMVC}. Since then there has been a steady progress on using attention primitives within vision models for recognition and classification \cite{Hu_2019_ICCV, Wang_2018_CVPR, Bello_2019_ICCV, Wang_2017_CVPR, ramachandran2019standalone, Zhao_2020_CVPR, Dosovitskiy_2020_ArXiv, Vaswani_2021_ArXiv, Srinivas_2021_arxiv}, detection \cite{Carion_2020_ECCV, Zhu_2021_ICLR}, segmentation \cite{Wang_2020_ECCV, Huang_2019_ICCV},  tracking \cite{Yu_2020_CVPR} and video analysis \cite{Pramono_2019_ICCV, Oh_2020_PAMI}. There have been numerous works interpreting the attention mechanism as a form of computing non-local means \cite{Buades_CVPR_2005, Wang_2018_CVPR}, approximating dynamic convolution filters \cite{Li_2021_CVPR, Vaswani_2021_ArXiv}, and capturing global context through long range interactions \cite{Hu_2019_ICCV, ramachandran2019standalone, Wang_2020_ECCV}. The standard dot-product attention \cite{Vaswani_2017_NIPS} update was also formulated as emerging from the update rule of modern Hopfield networks \cite{Ramsauer_2021_ICLR}. Our work introduces attention mechanism from a novel perspective as that of inferring from a probabilistic memory bank. To our knowledge, the only work that is closest to our approach is \cite{Ding_2020_ArXiv}, which also proposes a similar interpretation but only for the queries in order to study the explaining away effect of attention update. Our model encapsulates queries and values in a single generative model and provides an interpretation of standard dot-product attention as constrained Bayesian 
inference. Doubly normalized attention scheme \cite{Ding_2020_ArXiv} also emerges as a special case of key adaptation in our framework.       

\subsection{Interactive Segmentation}
Deep neural networks have set state-of-the art in semantic segmentation through the use of fully convolutional architectures \cite{Long_2015_CVPR,Zhao_2017_CVPR, Chen_2018_PAMI, Chen_2017_CVPR,Chen_2018_ECCV, Sun_2019_CVPR, Wang_2019_PAMI, Yuan_2020_ECCV} and more recently using  hybrid convolution and self-attention \cite{Yin_2020_ECCV} or stand-alone self-attention architectures \cite{Wang_2020_ECCV}. The input domain of interactive segmentation includes user input in the form of clicks or scribbles in addition to the visual signal (images or videos). The earliest works in interactive segmentation were based on algorithmic approaches for incorporating human inputs into region \cite{Rother_2004_ATG, Tang_2013_ICCV, Yu_2017_ICIP} or boundary \cite{Mortensen_1998_GMIP, Falcao_1998_GMIP} processing pipelines. \cite{Ramadan_2020_CVM} provides a comprehensive survey of interactive segmentation approaches. More recently deep networks have been used to incorporate user feedback to guide their output predictions at the pixel level. Following a similar taxonomy, these can be roughly categorized into region based \cite{Xu_2016_CVPR, Xu_2017_BMVC,Mahadevan_2018_BMVC, Li_2018_CVPR,Benenson_2019_CVPR,Agustsson_2019_CVPR,Hu_2019_NN,Zhang_2020_CVPR,Lin_2020_CVPR,Kontogianni_2020_ECCV} or boundary based approaches \cite{Castrejon_2017_CVPR,Acuna_2018_CVPR,Ling_2019_CVPR,Wang_2019_CVPR}. Deep Extreme Cut (DEXTR) \cite{Maninis_2018_CVPR} demonstrated that user guidance in the form of extreme points could be used in addition to the input channels to accurately localize the object of interest. More recently \cite{Zhang_2020_CVPR} argued that three points are sufficient as input guidance to localize the object but additional corrective clicks could be used to further refine the prediction. Other works have used the corrective clicks to adapt the network inputs \cite{Jang_2019_CVPR}, embeddings \cite{Sofiiuk_2020_CVPR} or parameters \cite{Kontogianni_2020_ECCV} online. Different from these previous approaches, we use corrective clicks as providing fixed values for a subset of units in the proposed probabilistic attention framework. These values are propagated globally through the attention mechanism to directly and more effectively influence the outputs towards user intended values.       

\section{Method}


We provide a probabilistic interpretation of attention as a generative model for queries and values through 
a set of memory units. Using this formulation, traditional attention in transformers \cite{Vaswani_2017_NIPS} reduces to the special case of maximum a posteriori (MAP) 
inference of values given queries, assuming Gaussians for the likelihoods. 
Using Bayesian inference, we provide a systematic approach to adapt keys online
as a locally ML update of the corresponding model parameters. Our formulation also allows to fix the values of certain units and propagate their influence to other units 
online by conditioning on the fixed unit values. 
The following sections provide more details on the probabilistic model.             
\subsection{Probabilistic attention}\label{sec:prob_attention}
We 
assume that there are $n$ memory units, indexed by $i$, each of which can be queried through a vector $q_i \in R^d$ to yield an output value vector $v_i \in R^m$. The queries and the corresponding values may depend on an input $x$. For example, each memory unit may represent a pixel $x_i$ in an image $x$. The joint distribution of queries $q_i$ and values $v_i$ conditioned on the input $x$ is assumed to factorize over memory units 
\begin{align}
p(q_{1:n}, v_{1:n} | x) = \Pi_{i=1}^n p_i(q_i,v_i \given x ),
\end{align}
where $x$ is the conditioning input, $q_{1:n} = \{ q_1, \cdots, q_n\}$, $v_{1:n} = \{v_1,\cdots, v_n\}$, and   $q_i \in R^d, v_i \in R^m$ are the query, value vectors for unit $i$ respectively. The per-unit joint likelihood $p_i(q_i,v_i \given x )$ is a probabilistic mixture model given by
\begin{align}
p_i(q,v \given x ) = \sum_{j=1}^n p_i(q,v,u_j\given x ) 
= \sum_{j=1}^n \pi_{i,j} (x) \:  p_i(q,v\given u_j,x) , 
\end{align}
where we have dropped the subscript $i$ from $q$ and $v$ for simplicity. In the above, $u_j$ indexes unit $j$, $\pi_{i,j}(x)$ is the probability of activating unit $j$ when unit $i$ is queried, $p_i(q,v \given u_j,x )$ is the likelihood of observing the pair $(q,v)$, given the pair is generated through unit $j$ in the mixture, conditioned on the input $x$. 

\subsection{Value inference}\label{sec:map_value_inference}
Using the above
model, it is possible to find the most 
likely value $\hat{v}$ given a query $q$ to unit $i$

\begin{align}
\hat v = \argmax_v p_i(v| q,x) .
\end{align}
We use Expectation Maximization (EM) \cite{Dempster_1977_JRSS} to achieve this: starting with an initial estimate $v^0$ of the most probable value and iterating over the standard EM auxiliary function $Q_i$. Given the latest  known estimate $v^t$, the $M$ step produces a new estimate $v^{t+1}$ that increases $Q_i$ by maximizing it \wrt $v^{t+1}$. 
This guarantees local maximization of $p_i(v|q,x)$. 
\begin{align}
&Q_i(v^t,v^{t+1}\given x) = \sum_j w_{i,j}^t \log p_i(u_j,q,v^{t+1}\given x),
\label{eqn:Q_function}
\end{align}
where
\begin{align}
w_{i,j}^t = p_i(u_j\given q,v^t, x) =  \frac{ \pi_{i,j}(x)  \:p_i(q,v^t \given u_j, x) }{\sum_j  \pi_{i,j}(x)  \:p_i(q, v^t\given u_j, x) }.
\end{align}
The $n\times n$ matrix formed by the entries $w_{i, j}$ corresponds to the {\em attention} matrix in standard transformers. 
The optimal value 
$\hat{v}$ is obtained by  taking the gradient with respect to $v^{t+1}$ and setting it to zero 
\begin{align}\label{eqn:general}
	\nabla_{v^{t+1}} Q_i(v^t,v^{t+1} \given x ) = \sum_j w_{i,j} \nabla_{v^{t+1}} \log p_i(q, v^{t+1}, u_j \given x), 
\end{align}
where  $\nabla_{v^{t+1}}  \log p_i(q, v_{t+1}, u_j \given x)$ is the Fisher Score for unit $i$ with respect to $v^{t+1}$.

\subsection{Relationship to standard attention}\label{sec:relate_std_attention}
We show that standard attention in transformers \cite{Vaswani_2017_NIPS} solves Eq.~\eqref{eqn:general} under the special case of a constrained Gaussian mixture model (GMM). Assuming 
isotropic Gaussians with conditionally independent queries and values given input and mixture component
\begin{align}
 &p_i(q,v \given u_j,x)  = p_i(q\given u_j,x) \: p_i(v \given u_j,x) \\
&p_i(q \given u_j,x) = \Big( \frac{\alpha_j(x)}{2 \pi}\Big)^{d/2}\; e^{-\frac{\alpha_j(x)}{2}  \| q - \xi_j(x)\|^2 }\\
&p_i(v \given u_j,x) = \Big( \frac{\beta_j(x)}{2 \pi}\Big)^{m/2}\; e^{-\frac{\beta_j(x)}{2}  \| v - \mu_j(x)\|^2 },
\end{align}
 where $\alpha_j(x), \beta_j(x) > 0$ are precision parameters,  $\xi_j(x) \in R^d$, $\mu_j(x) \in R^m$ are the key and expected value parameters for unit $j$ given the input $x$.  
 The dependency of $p_i(q,v \given u_j,x)$ on $x$ is through the fact that the parameters $\alpha_j, \beta_j, \pi_{i,j}$, $\xi_j, \mu_j$ are a function of $x$. For simplicity, we treat $x$ to be fixed and leave the dependency on $x$ implicit in our notation. 
In order to obtain the standard attention update equation, we constrain the precision parameters to be the same across units: $\alpha_1 = \cdots = \alpha_n = \alpha$,  $\beta_1 = \cdots = \beta_n = \beta$, and link the priors of each unit to the lengths of the corresponding key and expected value vectors
\begin{align} \label{eqn:mag_prior}
\pi_{i,j}  &=\frac{1}{z} e^{\frac{\alpha}{2} \| \xi_j\|^2 }  e^{\frac{\beta}{2} \| \mu_j\|^2 } \\
z &= \sum_j e^{\frac{\alpha}{2}  \| \xi_j\|^2 }  e^{\frac{\beta}{2} \| \mu_j\|^2 }  .
\end{align}
Assuming $\beta \to 0$ and solving for optimal $v^{t+1}$ in Eq.~\eqref{eqn:general},  we obtain the standard attention update (see Appendix~\ref{appx:relation_std_attention})
\begin{align}\label{eqn:optimalvstdatt}
&v^{t+1} = \sum_j w_{i,j} \mu_j\\
&w_{i,j}^t = \frac{ e^{\alpha \xi_j^T q}   }
{\sum_j e^{\alpha\xi_j^T q}} ,
\end{align}
where each $\mu_j$ is the value associated with unit $j$ and $v^{t+1}$ is the output at unit or token $i$ after the attention update. 
In this case, $w_{i,j}^t$ is no longer a function of $t$ and thus only one EM iteration is needed. 

\subsection{Offline supervised learning}\label{sec:offline_supervised_learning}
As is commonly done in standard transformers, the relationship between the input $x$ and the mixture model  parameters: $\pi(x), \xi(x), \mu(x)$ can be modeled using a deep network, whose parameters can be trained off-line with task specific supervision. 

\subsection{Online unsupervised mixture model adaptation}\label{sec:unsupervised_model_adaptation}
Our framework provides a way to adapt the mixture model parameters based on all the observed input queries prior to doing 
value inference. This process can be seen as an inference-time adaptation of the model using the additional information contained in the set of queries. We propose an unsupervised Bayesian approach to do this adaptation for the per-unit key vectors $\xi_{1:n} = \{ \xi_1,\cdots, \xi_n\}$ and precision parameters $\alpha_{1:n} = \{ \alpha_1,\cdots, \alpha_n\}$ given queries $q_{1:n} = \{ q_1, \cdots, q_n\}$. For each unit $i$, the optimal value inference is given by  
\begin{align}
&\hat v_i = \argmax_v p_i (v \given q_{1:n})  .
\end{align}
Assuming a prior for the key vectors given the observed queries $p(\xi_{1:n}  \given q_{1:n} )$, the likelihood $p_i (v \given q_{1:n})$ can be written as
\begin{align}
p_i( v \given  q_{1:n} ) = \int p(\xi_{1:n}  \given q_{1:n} ) p_i (v \given q_i, \xi_{1:n}) d \xi_{1:n} 
\approx  p_i( v \given q_i,\hat \xi_{1:n}) ,
\end{align}
where the expectation over the posterior is approximated by its maximum a posteriori (MAP) value
\begin{align}
&\hat \xi_{1:n} = \argmax_{ \xi_{1:n}}   p (\xi_{1:n}  \given q_{1:n})\label{eqn:maxxi}\\
&\hat v_i = \argmax_v p_i (v \given q_i, \hat  \xi_{1:n}  ) . \label{eqn:maxv1}
\end{align}
In order to solve \eqref{eqn:maxxi}, we use an iterative EM approach. The initial key parameters $\xi^0_{1:n}$ are provided by the pre-trained model. To avoid overfitting to the current query vectors, we use a Gaussian prior centered on the key parameters provided by the pre-trained network, i.e., $\xi^0_{1:n}$ with a finite precision $\theta_{\xi} >0 $. 
The EM update for the key parameters at any iteration $t$ is given by (see Appendix~\ref{appx:online_key_adaptation}) 
\begin{align}
\xi_k^{t+1} = \frac{\theta_{\xi} \xi^t_k  + \alpha_k \sum_{i=1}^n w_{i,k}^t q_i }{ \theta_\xi + \alpha_k \sum_{i=1}^n w_{i,k}^t} .
\label{eqn:key_adaptation}
\end{align}
Analogous to the keys, we can also adapt the $\alpha_j$ precision parameters (see Appendix~\ref{appx:online_alpha_adaptation}). 

\subsection{Online value propagation}\label{sec:online_value_propagation}
The proposed model allows for fixing the outputs of a selected subset of units to predefined values and letting them propagate to other units in a principled way. This aspect of our model is of particular interest to interactive semantic segmentation, where a human annotator provides corrections to the output of a semantic segmentation model. In this case, assuming an attention layer at the output of a deep model, the memory units correspond to pixels and the output values correspond to the semantic label for that pixel, \eg foreground or background. Based on the network's prediction, an annotator provides corrections for a subset of the pixels, which are the ground truth for those pixels. These correspond to the fixed predefined values for those units, whose effect is to be propagated to semantically similar pixels globally across the image to make the process more efficient. More formally, suppose  the annotator has provided the correct values  for the first $s<n$ units.  We want for this information to improve the inference about the value for all the other units $i > s$.  Within our framework, this inference is given by         
\begin{align}
\hat v_i = \argmax_v p_i (v \given q_i, q_{1:n}, v_{1:s} ), \: \text{for $s<i<=n$} .
\end{align}
In order to do this inference, we adopt a 
Bayesian approach similar to model adaptation of Section~\ref{sec:unsupervised_model_adaptation}. Let $\lambda$ represent the set of network parameters, e.g., $\pi,\xi,\mu, \alpha, \beta$. Writing the inference as an expectation over the model posterior $p(\lambda\given q_{1:n}, v_{1:s}  )$
\begin{align}
p_i( v \given  q_{1:n}, v_{1:s} ) = \int p(\lambda\given q_{1:n}, v_{1:s}  ) p_i(v \given q_i, \lambda) d \lambda 
\approx  p_i( v \given q_i,\hat \lambda),
\end{align}
where we approximate the expectation with its MAP estimate as before
\begin{align}
&\hat \lambda = \argmax_\lambda p (\lambda \given q_{1:n}, v_{1:s})\label{eqn:lambda}\\
&\hat v_i = \argmax_v p_i(v \given q_i, \hat \lambda).
\end{align}
Eq.~\eqref{eqn:lambda} is solved using EM. Specifically, value propagation across units is achieved by updating the $\mu_k$ for each unit $k$ starting with the initial value $\mu_k^0$ provided by the pre-trained model (Section~\ref{sec:offline_supervised_learning}). Following a similar approach as in Section~\ref{sec:unsupervised_model_adaptation}, the EM update for $\mu_k^{t+1}$ at iteration $t$ is given by
\begin{align}
&\mu_k^{t+1} = \frac{\theta_{\mu} \mu^t_k  + \beta_k \sum_{i=1}^s w_{i,k}^t v_i }{ \theta_\mu+ \beta_k \sum_{i=1}^s w_{i,k}^t}
\label{eqn:value_propagation}
\end{align}
\begin{align}
w_{i,k}^t = p_i(u_k \given q_i,v_i, \mu^t_{1:n} )  
= \frac{\pi_{i,k} \: p(q_i \given u_k, \xi_k )\:p(v_i \given u_k, \mu^t_k )}{\sum_{j=1}^n \pi_{i,k} \: p(q_i \given u_j, \xi_j )\:p(v_i \given u_j, \mu_j^t )},
\label{eqn:value_propagation_wts}
\end{align}
where $\theta_\mu$ is the precision for the Gaussian prior over each $\mu_k$. See also Appendix~\ref{appx:value_params_update}.

\subsection{Combining offline learning and online adaptation} 
The inference time  adaptation of parameters is differentiable. So it can be included as part of the traditional supervised optimization \eg via stochastic gradient descent and used to learn the parameters of the prior distributions over $\xi,\mu, \alpha, \beta,\pi$.  

\subsection{Position embeddings}\label{sec:position_embeddings}
Positional embeddings \cite{Vaswani_2017_NIPS, Shaw_2018_NAACL} are useful in attention models to encode the relative or absolute positions of tokens. In computer vision applications, relative position embeddings have been found to be critical to capture the interactions between features based on their pairwise positional relations \cite{ramachandran2019standalone, Srinivas_2021_arxiv, Wang_2020_ECCV}. We propose to encode relative position embeddings by introducing extra parameters in the per-unit likelihoods of the mixture components and their priors. Let $r_{j-i}^q$, and $r_{j-i}^k$ denote the relative position embeddings for a query and key interacting at units $i$ and $j$ respectively. The query/key marginal with the position embeddings is given by (see Appendix~\ref{appx:query_lhood_with_position_embeddings})
\begin{align}
    p_i(q \given \xi_j, r_{j-i}^q, u_j) \propto
    \mathcal{N}(q \given \xi_j, \, \frac{1}{\alpha_j}I_d) \mathcal{N}(q \given r_{j-i}^q, \, \frac{1}{\alpha_j}I_d) 
    \propto \mathcal{N}(q \given \frac{\xi_j+r_{j-i}^q}{2}, \, \frac{1}{2\alpha_j}I_d), 
    \label{eqn:query_lhood_with_position_embeddings}
\end{align}
where $\mathcal{N}(a \given b, \, c)$ is the Gaussian likelihood function over $a$ with mean $b$ and 
covariance matrix $c$. $I_d$ is a $d \times d$ identity matrix. The mixture component priors with position embeddings take the form
\begin{align}
\pi_{i,j} \propto \mathcal{N}(\xi_j \given r_{j-i}^k, \, \frac{1}{\alpha_j}I_d) 
\exp{\left[\frac{\alpha_j}{2} \left(2\|\xi_j\|^2 + \|r_{j-i}^q\|^2 + \|r_{j-i}^k\|^2 \right)\right]} \exp{\left[\frac{\beta_j}{2} \| \mu_j\|^2\right]} .
\end{align}

\section{Experiments}
In this section, we report the results of using probabilistic attention at various stages of a deep interactive semantic segmentation network. Specifically, we use it within the BoTNet50 backbone \cite{Srinivas_2021_arxiv} in place of standard attention and also as part of a self-attention based classification head at the network output. We quantify model performance (mean IOU relative to ground truth) as a function of the number of clicks \cite{Benenson_2019_CVPR} on two widely used public benchmarks for this task: GrabCut\cite{Rother_2004_ATG} and Berkeley \cite{Martin_2001_ICCV}. Appendix~\ref{appx:models_training_evaluation} provides more details on the interactive segmentation model architectures, training and evaluation protocols.

\subsection{Probabilistic attention within a backbone}\label{sec:probabilistic_attention_backbone}
\begin{figure*}
	\centering
	\subfloat[GrabCut]
	{
			\includegraphics[width=0.5\linewidth]{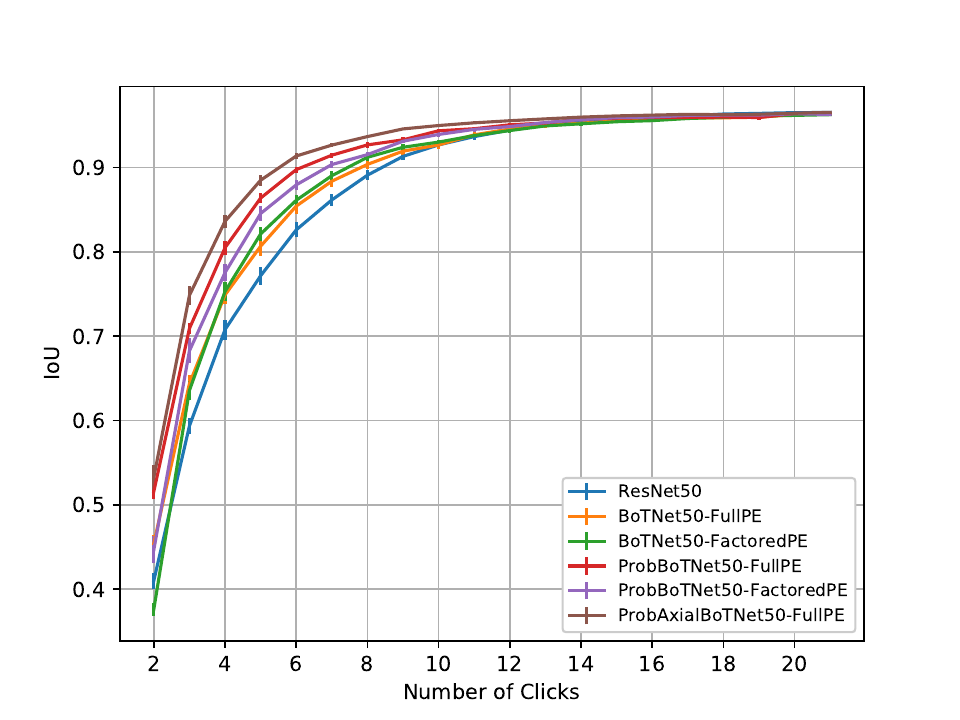}
			\label{fig:probatt_botnetbb_grabcut}
	}
	\subfloat[Berkeley]
	{
			\includegraphics[width=0.5\linewidth]{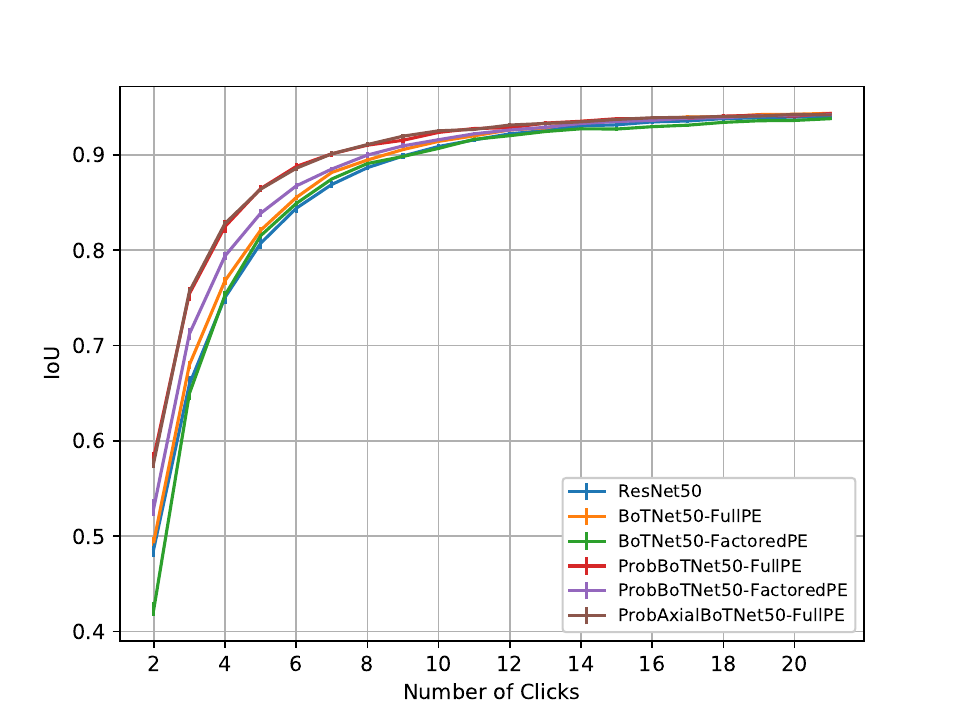}
			\label{fig:probatt_botnetbb_berkeley}
	}
	\caption{\textbf{Probabilistic attention layers in BoTNet architecture}. Mean IoU as a function of clicks using different attention layers, position embeddings and full or axial attention in the BoTNet architecture.  These are compared against their fully convolutional conterpart ResNet50.
	The left and the right plots correspond to the GrabCut and Berkeley datasets respectively.}
    \label{fig:probatt_botnetbb}
\end{figure*}

We adopt the recent work on BoTNet \cite{Srinivas_2021_arxiv} by replacing the convolutional layers with attention layers in the last bottleneck block (c5) of the ResNet50 \cite{He_2016_CVPR} architecture. Specifically, we use probabilistic attention layers in place of standard attention using either full or axial \cite{Wang_2020_ECCV} attention. We experiment with either factored \cite{Srinivas_2021_arxiv} or full relative positional encoding. Factored encoding uses $(H+W)d$ parameters for an image of size $(H, W)$ factoring them along the height and width dimensions, whereas full encoding uses $2(H*W)d - d$ parameters, $d$ per relative offset. Our models are trained on the LVIS \cite{Gupta_2019_CVPR} dataset at a resolution of 256 pixels. The results are shown in Fig.~\ref{fig:probatt_botnetbb}.  
The results suggest that using probabilistic attention in the BoTNet50 backbone leads to better performance especially for smaller number of clicks. This is true with both full and axial attention BoTNets using probabilistic attention. Using full relative position encoding helps more than using factored encoding perhaps due to the larger number of parameters.    

\subsection{Key adaptation}\label{sec:key_adaptation}
\begin{figure*}[t!]
    \centering
    \subfloat[GrabCut]
    {
        \includegraphics[width=0.5\linewidth]{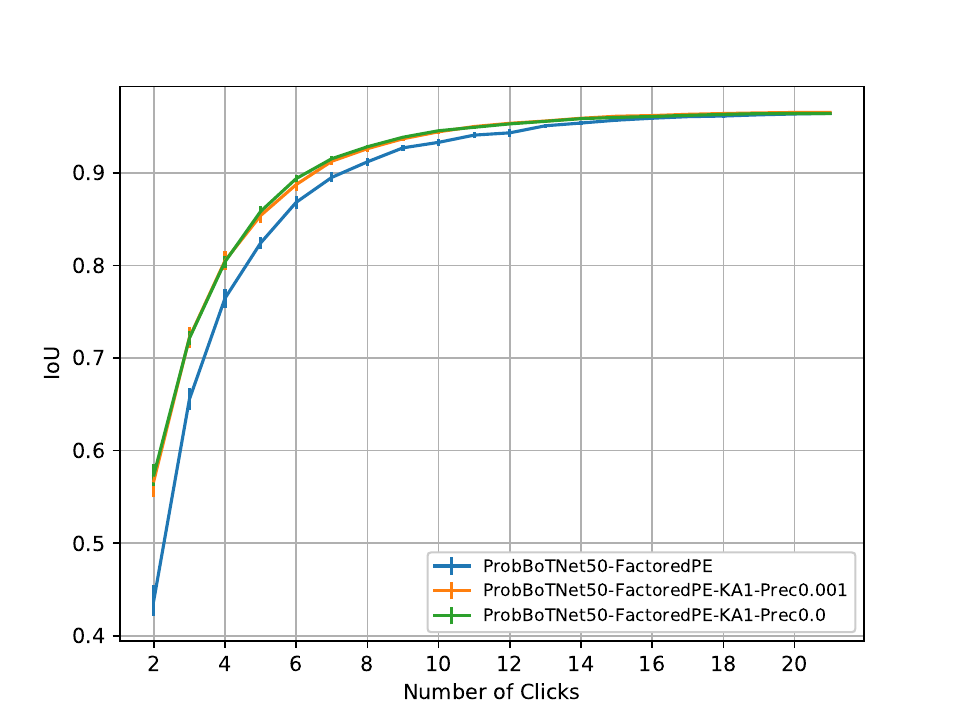}
        \label{fig:probatt_botnetbb_factoredPE_KA_grabcut}
    }
    \subfloat[Berkeley]
    {
        \includegraphics[width=0.5\linewidth]{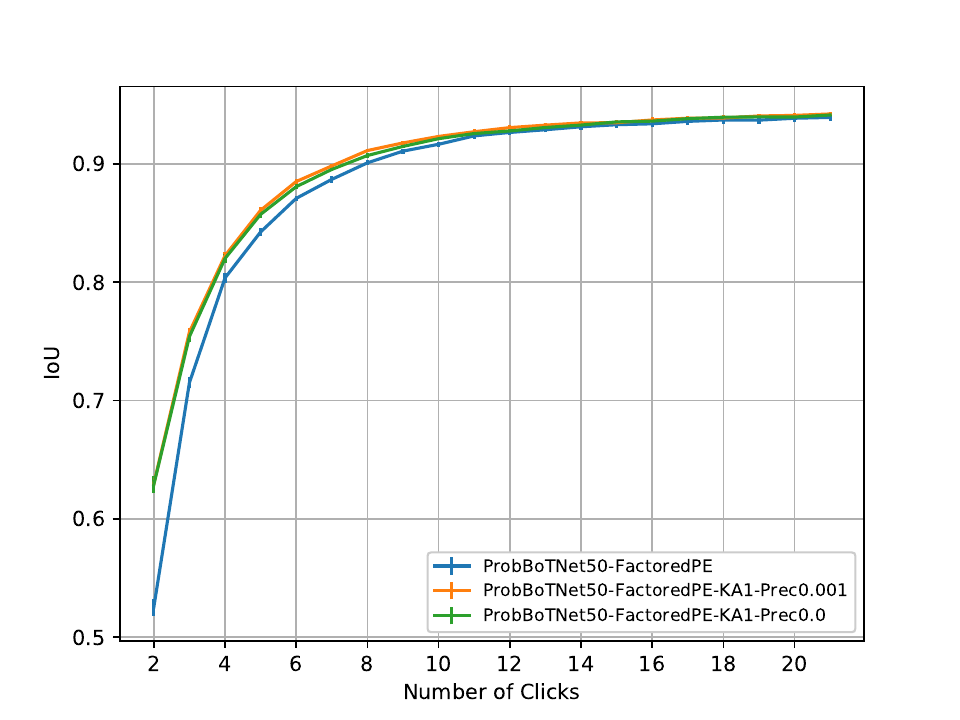}
        \label{fig:probatt_botnetbb_factoredPE_KA_berkeley}
    } 
    \caption{\textbf{Unsupervised key adaptation}. Mean IoU vs \#clicks with and without key adaptation (KA) on the GrabCut and Berkeley datasets. Probabilistic BoTNets with 
    factored position encodings are evaluated without using KA or using 1 iteration of KA with two different prior precision (Prec.) values of 0.001 or 0.   
    }
    \label{fig:probatt_botnetbb_KA}
\end{figure*}

We experiment with unsupervised model adaptation as described in Section~\ref{sec:unsupervised_model_adaptation} by adapting the keys online (Eq.~\eqref{eqn:key_adaptation}) based on the observed queries. The degree of adaptation is controlled by the prior precision parameter $\theta_{\xi}$ with lower values leading to a higher degree of adaptation due to the lower weight on the prior keys. Using the probabilistic attention BoTNet50 backbones of the previous section, we experiment with and without key adaptation. With key adaptation, we use two different values of the precision prior, 0.001 and 0, with the latter corresponding to a maximum likelihood update of the keys given observed queries. The results in Fig.~\ref{fig:probatt_botnetbb_KA} show the mean IoU as a function of number of clicks using the ProbBoTNet50-FactoredPE model.
We observe that key adaptation leads to higher IoUs without any corrective clicks or using only a few corrective clicks. Specifically there is an absolute improvement of about 10\% in mean IoU using key adaptation and without using any corrective clicks. 
Additional results using ProbBoTNet50-FullPE are shown in the the Appendix~\ref{appx:key_adaptation_results}. Using a lower value of prior precision seems beneficial and the extreme case of maximum likelihood adaptation leads to the best performance. Note that this effect has been observed in a previous work \cite{Ding_2020_ArXiv}, where it is perceived as a doubly normalized attention scheme (DNAS). This can be attributed to the unsupervised model adaptation accounting for the small domain shift introduced by 
models trained on LVIS and evaluated on GrabCut and Berkeley datasets. Without using key adaptation additional user input in the form of corrective clicks is required to account for this shift as can be seen by the asymptotically similar behavior with increasing number of clicks.

\subsection{Probabilistic attention as a classifier}\label{sec:probabilistic_attention_classifier}
Deep architectures \cite{Long_2015_CVPR,He_2017_ICCV,Chen_2018_ECCV, Chen_2018_PAMI} for semantic segmentation have a fully connected (1x1 conv) layer as their classification head. We replace this layer with a corrective self-attention \cite{Vaswani_2017_NIPS} module that takes corrective click locations as additional inputs to more effectively propagate corrections as follows. 

\begin{figure*}[!htbp]
   \includegraphics[width=1.0\linewidth]{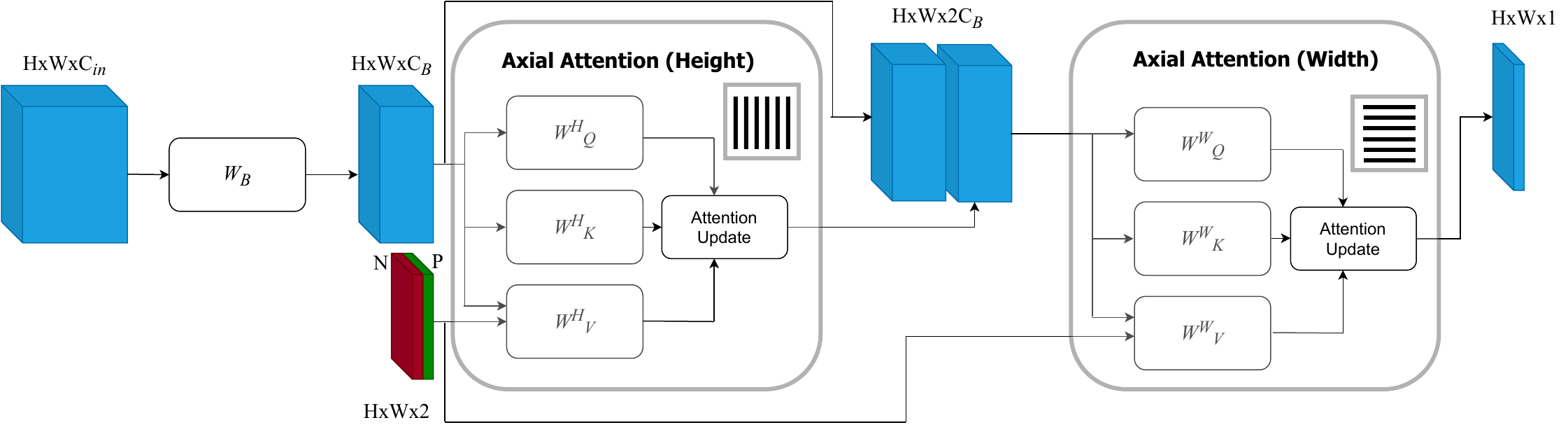}
   \caption{\textbf{Corrective self attention layer}. We propose a self-attention based classification head at the output of the decoder to more effectively propagate corrective clicks. $C_{in}$ channels from the penultimate decoder layer are reduced to $C_B$ by a bottleneck layer with weights $W_B$. These are input to a pair of densely connected Axial Attention \cite{Wang_2020_ECCV} modules along height and width dimensions to produce the output logit at full image resolution ($H \times W$). The corrective channels ($P$ and $N$) are fed only to the value embedding functions ($W^H_V$ and $W^W_V$) of the attention modules to propagate the corrections more effectively.} 
   \label{fig:CSA}
\end{figure*}

\textbf{Corrective self attention.} We append the corrective channels to the features of the penultimate decoder layer and feed them as inputs to the value embedding layer (see Fig.~\ref{fig:CSA}). The query and the key embeddings do not use the corrective channels as their inputs, which allows attention maps to be computed based only on the semantics captured by the features. However, the weights of the value embedding layer can be trained to output the desired labels at the locations of the corrective clicks.
Fig.~\ref{fig:CSA} shows a block diagram of our corrective self-attention (CSA) layer. We choose to use probabilistic self-attention for computing the output values. 

The CSA layer is used at the network output by up-sampling the final feature map of the decoder to the input image resolution and appending positive and negative corrective channels 
containing only the click locations. The local context size for Axial attention modules is chosen to be 64 pixels.
The output of the axial attention block is passed through a sigmoid to estimate the pixelwise probabilities of the object mask.

\subsection{Value propagation}\label{sec:value_propagation_results}
\begin{figure*}
	\centering
	\subfloat[GrabCut]
	{
			\includegraphics[width=0.5\linewidth]{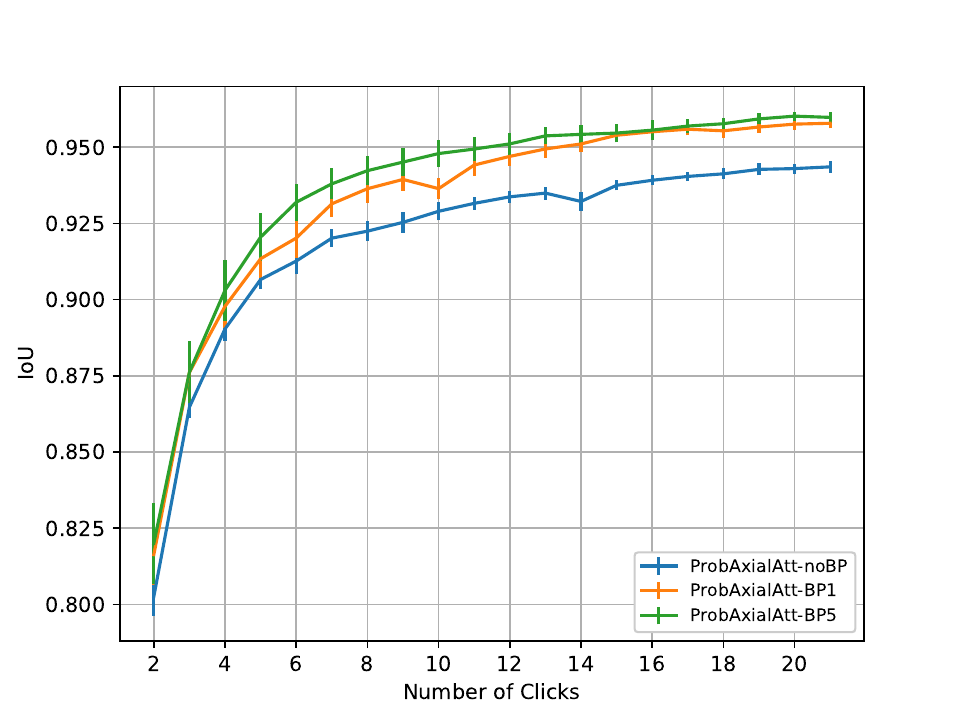}
			\label{fig:probfaxialatt_BP_grabcut}
	}
	\subfloat[Berkeley]
	{
			\includegraphics[width=0.5\linewidth]{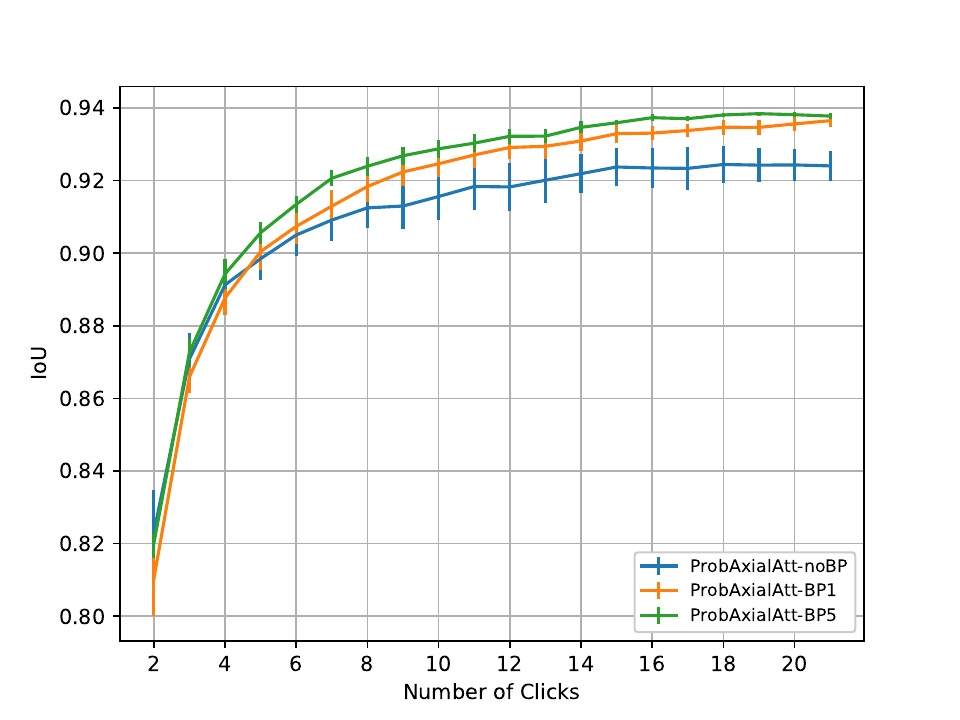}
			\label{fig:probaxialatt_BP_berkeley}
	}
	\caption{\textbf{Effect of value propagation using Axial attention}. Mean IoU vs \#clicks with and without value propagation using an axial attention based probabilistic CSA layer at the output.
	We use 1 (BP1) and 5 (BP5) iterations of value propagation at the CSA layer and test on the GrabCut (left) and Berkeley (right) datasets.
	}
    \label{fig:probaxialatt_BP}
\end{figure*}

We 
demonstrate the effect of propagating annotator feedback across pixels using online value propagation as described in Section~\ref{sec:online_value_propagation}. 
We use the CSA layer described above in place of the 1x1 conv classifier head of the HRNetV2+OCR \cite{Yuan_2020_ECCV} architecture pre-trained on Imagenet classification and fine-tuned for interactive segmentation on SBD \cite{Bharath_2011_ICCV} at a resolution of 256 pixels.
Note that value propagation requires learning one additional parameter per output class (2 in our case) to estimate the fixed logit that corresponds to the annotator feedback for that class at the corrective locations. These are learnt as part of network training using gradient descent. 
Following standard protocols, we test on the GrabCut \cite{Rother_2004_ATG} and Berkeley \cite{Martin_2001_ICCV} datasets.
Fig.~\ref{fig:probaxialatt_BP} shows the effect of using different number of value propagation iterations (1 and 5) within the probabilistic attention layer. Clearly, value propagation leads to more effective propagation of labeler feedback relative to not using it.   
For this experiment, we do value propagation only in the output width block of the CSA layer (Fig.~\ref{fig:CSA}) as we found that doing so in both the height and width layers did not work so well in our experiments. We hypothesize that this is probably due to the difficulty in learning the high dimensional fixed parameters corresponding to annotator feedback in the height block of the axial attention layer. 

\begin{figure*}
	\centering
	\subfloat[GrabCut]
	{
			\includegraphics[width=0.5\linewidth]{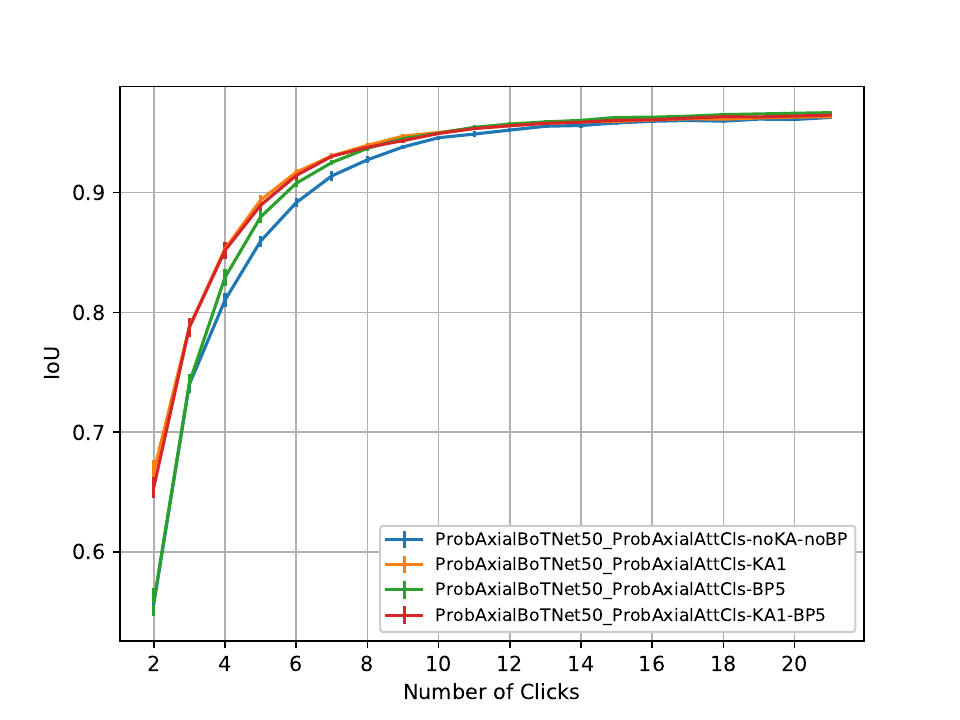}
			\label{fig:probaxialatt_KABP_grabcut}
	}
	\subfloat[Berkeley]
	{
			\includegraphics[width=0.5\linewidth]{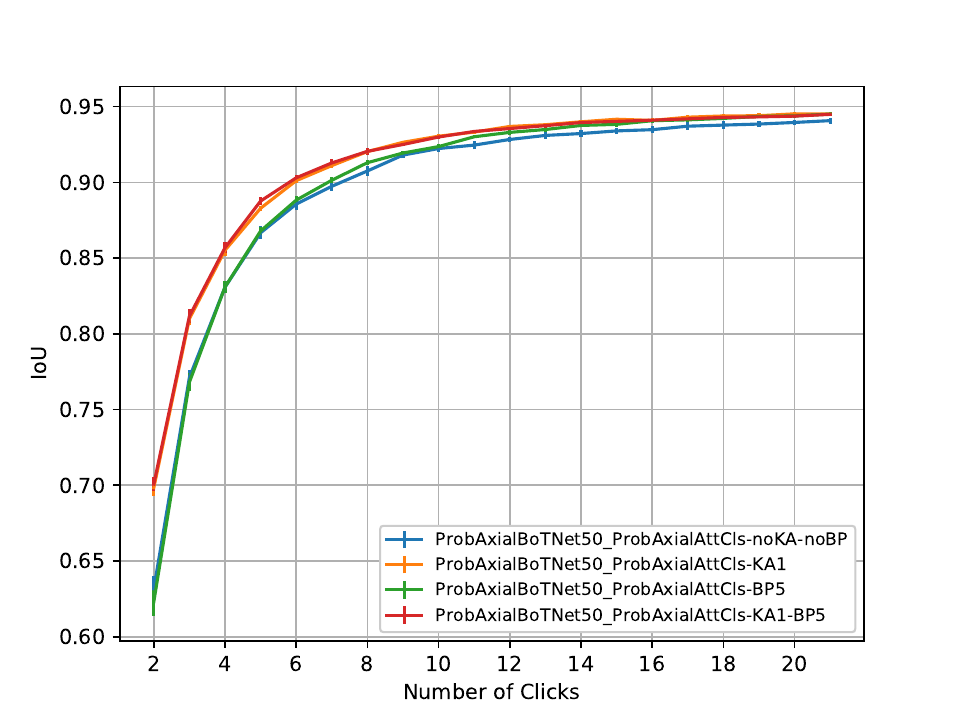}
			\label{fig:probaxialatt_KABP_berkeley}
	}
	\caption{\textbf{Effect of combining key adaptation and value propagation}. Mean IoU vs. \#clicks using one or both of key adaptation and value propagation in a single model. Key adaptation is run for 1 iteration (KA1) and value propagation for 5 iterations (BP5) when either or both are used.}
\label{fig:probaxialatt_KABP}
\end{figure*}

\subsection{Combining key adaptation and value propagation}
In this section, we experiment with combining key adaptation and value propagation in a single model. For this experiment we use a BoTNet50 architecture with a corrective self-attention classification head at a resolution of 256 pixels. We use axial attention in both the backbone and the classification head with probabilistic self attention updates. The model is trained on LVIS and evaluated on GrabCut and Berkeley datasets. Fig.~\ref{fig:probaxialatt_KABP} shows the effect of using either key adaptation or value propagation or both relative to not using them. As observed separately in the previous plots, key adaptation helps in the small \#clicks regime whereas value propagation shows greater benefits with increasing \#clicks. Using the two jointly allows the model to respond quickly to annotator feedback in both the regimes.  

\section{Conclusion}
We provide a probabilistic interpretation of the attention mechanism in transformers as a generative mixture model over queries and values parameterized through keys. Using our framework, the attention update is 
maximum a posteriori inference over values given queries. Specifically, the standard dot-product attention is a special case assuming Gaussians for the mixture likelihoods and a few other constraints. Using Bayesian inference, our interpretation allows for online update of the mixture model parameters as well as the propagation of a set of fixed values specified by an external agent. Although we demonstrate the utility of these aspects on the problem of interactive segmentation, the proposed model is generic and can be extended to other domains with suitable distributional forms for the mixture likelihoods.  

\bibliographystyle{ieee_fullname}
\bibliography{probabilistic_attention}
\appendix


\section{Graphical model}
\begin{figure}[htbp]
    \centering
    \includegraphics[width=0.25\linewidth]{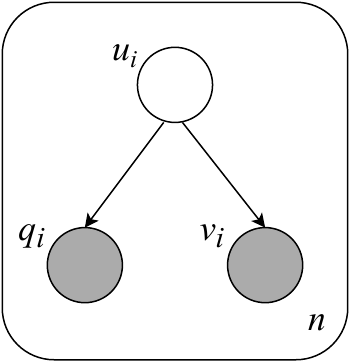}
    \caption{\textbf{Probabilistic generative model for queries and values}. Graphical representation of the generative model for a query ($q_i$) and a value ($v_i$) through a corresponding hidden latent variable ($u_i$) that indexes over units of a probabilistic memory bank. $n$ denotes the \#units in the memory bank as well as the number of generated query/value pairs.}
    \label{fig:probatt_graph}
\end{figure}

The constrained generative model equivalent to standard dot-product attention (Section~\ref{sec:relate_std_attention}) can be expressed as a Bayesian probabilistic graphical model shown in Fig~\ref{fig:probatt_graph}. In order to generate a query $q_i$ (observed) and a value $v_i$ (observed) at unit $i$, a memory unit $u_i$ (unobserved) is first sampled from a prior $\pi_{ij}$ over units $j$ in the the memory bank. This is done independently for each unit across the memory bank comprising of $n$ total units. The per-unit queries and values are then sampled independently from isotropic Gaussians as described in Section~\ref{sec:relate_std_attention}.

\section{Proofs}
\subsection{Relationship to standard attention}\label{appx:relation_std_attention}
We provide a detailed proof of Eq.~\eqref{eqn:optimalvstdatt}. With the assumptions of Section~\ref{sec:relate_std_attention}, 
Eq.~\eqref{eqn:Q_function} is given by
\begin{align}
    Q_i(v^t, v^{t+1}) = \sum_j w_{i, j}^t \left(\frac{d}{2}\log \left(\frac{\alpha_j}{2\pi}\right) -\frac{\alpha_j}{2} \| q - \xi_j\|^2
    + \frac{m}{2}\log \left(\frac{\beta_j}{2\pi}\right)
    -\frac{\beta_j}{2}  \|  v^{t+1}  - \mu_j\|^2\right) ,
\end{align}
where $w_{i, j}^t$, including the precision parameter  $\beta_j$, is
\begin{align}
w_{i,j}^t  = \frac{ 
\pi_{i,j} \: \beta_j\:e^{-\frac{\alpha_j}{2}  \| q - \xi_j\|^2 } \:e^{-\frac{\beta_j}{2}  \| v^t - \mu_j\|^2 }}
{ 
\sum_j \pi_{i,j} \beta_j  \: e^{-\frac{\alpha_j}{2}  \| q - \xi_j\|^2 } \:e^{-\frac{\beta_j}{2}  \|  v^t  - \mu_j\|^2 }} .
\end{align}

Taking the derivative 
\wrt $v^{t+1}$ and setting it to zero,
 \begin{align}
	\nabla_{v^{t+1}} Q_i(v^t, v^{t+1}) = \sum_j w_{i,j}^t \beta_j  (\mu_j -v^{t+1}) =0,
\end{align}
the EM update equation reduces to
\begin{align}\label{eqn:gauss}
v^{t+1}  =  \sum_j  w_{i,j}^t\: \mu_j .
\end{align}

The prior of Eq.~\eqref{eqn:mag_prior} makes $w_{i,j}^t$ independent of $i$ (permutation equivariant) and simplifies the optimal value inference equation to
\begin{align}\label{eqn:optimalv}
&v^{t+1}  = \sum_j w_{i,j}^t \mu_j\\
&w_{i,j}^t = \frac{ e^{\alpha \xi_j^T q} \;e^{\beta   \mu_j^Tv_t }  }
{\sum_j e^{\alpha\xi_j^T q}\;  \;e^{\beta  \mu_j^T v_t }} .
\end{align}
It is easy to see that as $\beta \to 0$ we obtain the standard dot product attention update as in Eq.~\eqref{eqn:optimalvstdatt}.

\subsection{Online key adaptation}\label{appx:online_key_adaptation}
At any EM iteration $t$, the auxiliary function $Q(\xi_{1:n}^t , \xi_{1:n}^{t+1} )$ for key update is given by 
\begin{align}
Q(\xi_{1:n}^t , \xi_{1:n}^{t+1} ) &= \log p(\xi^{t+1}_{1:n}) + \sum_{i=1}^n \sum_{j=1}^n  w_{i,j}^t \log p_j( q_i, u_j \given \xi_j^{t+1}) ,
\end{align}
where
\begin{align}
&w_{i,j}^t = p_i(u_j \given q_i, \xi^t_{1:n} )  = \frac{\pi_{i,j} \: p(q_i \given u_j, \xi^t_j )}{\sum_{k=1}^n \pi_{i,k} \:  p(q_i\given u_k, \xi_k^t)}.
\end{align}
Taking the derivative \wrt the key vector $\xi_k^{t+1}$ of unit $k$  
\begin{align}
\nabla_{\xi_k^{t+1} } Q(\xi_{1:n}^t , \xi_{1:n}^{t+1} ) &= \theta_\xi ( \xi_k^t- \xi^{t+1}_k) + \sum_{i=1}^s w_{i,k}^t  \alpha_k (q_i - \xi^{t+1}_k ) .
\end{align}
Setting the derivative to zero and solving for $\xi_k^{t+1}$ leads to the online key adaptation update of Eq.\eqref{eqn:key_adaptation}.

\subsection{Online adaptation of $\alpha_j$ precision parameters}\label{appx:online_alpha_adaptation}
The precision parameters $\alpha_j$ in the per-unit query likelihoods can be adapted online based on the observed queries, similar to keys in Eq.~\eqref{eqn:key_adaptation}. In order to avoid overfitting to the observed queries, we use a Gamma prior with parameters $\theta_{\alpha,1}, \theta_{\alpha,2}$, which leads to the following update for the precisions
\begin{align}
\alpha_k^{t+1} = \frac{\theta_{\alpha,1}  + d/2 \sum_{i=1}^n w_{i,k}^t -1}{ \theta_{\alpha,2} +\sum_{i=1}^n w_{i,k}^t\: \frac{1}{2} \| q_i - \xi_k\|^2 } .
\end{align}

\subsection{Updates of value likelihood parameters based on fixed values}\label{appx:value_params_update}
In addition to value propagation (Eq.~\eqref{eqn:value_propagation}), the probabilistic attention model allows updating the per-unit value likelihood component parameters based on the information provided by the fixed pre-selected values. Specifically, the EM updates for the unit $k$ likelihood parameters $\beta_k$ and $\pi_{i,k}$ are given by
\begin{align}
&\beta_k^{t+1} = \frac{\theta_{\beta,1}  + d/2 \sum_{i=1}^s w_{i,k}^t -1}{ \theta_{\beta,2} +\frac{1}{2} \sum_{i=1}^s w_{i,k}^t \| v_i - \mu_k\|^2 }\\
&\pi_{i,k}^{t+1} = \frac{ w_{i,k}^t + \theta_{\pi,i,k} -1}{\sum_k w_{i,k}^t + \theta_{\pi,i,k} -1},
\end{align}
where $\theta_{\beta,1}, \theta_{\beta,2}$ are the parameters for a Gamma prior distribution over $\beta_k$, and $\theta_{\pi,i,k}$ are Dirichlet prior parameters over $\pi_{i,k}$. The weights $w_{i, k}^t$ above are the same as in Eq.~\eqref{eqn:value_propagation_wts}.

\subsection{Position embedding formulations}\label{appx:query_lhood_with_position_embeddings}
Here we provide details on how we arrive at the form of the per-unit query likelihood of Eq.~\eqref{eqn:query_lhood_with_position_embeddings}. By choosing to include the position embeddings $r_{j-i}^q$ through an extra normal likelihood function, the per-unit query likelihood is given by
\begin{align*}
    p_i(q \given \xi_j, r_{j-i}^q, u_j) \propto
    \mathcal{N}(q \given \xi_j, \, \frac{1}{\alpha_j}I_d) \mathcal{N}(q \given r_{j-i}^q, \, \frac{1}{\alpha_j}I_d), 
\end{align*}
where $\mathcal{N}(a \given b, c)$ is the Gaussian likelihood function over $a$ with mean $b$ and covariance matrix $c$. $I_d$ is a $d \times d$ identity matrix. Making use of the fact that the product of two normal likelihood functions is also a normal and completing the square
\begin{align}
    \mathcal{N}(q \given \xi_j, \, \frac{1}{\alpha_j}I_d) \mathcal{N}(q \given r_{j-i}^q, \, \frac{1}{\alpha_j}I_d)  
    = \mathcal{N}(q \given \frac{\xi_j+r_{j-i}^q}{2}, \, \frac{1}{2\alpha_j}I_d) \mathcal{N}(\xi_j \given r_{j-i}^q, \, \frac{2}{\alpha_j}I_d),
\end{align}
we arrive at the form in Eq.~\eqref{eqn:query_lhood_with_position_embeddings}. Note that there is effectively no direct interaction between $\xi_j$ and $r_{j-i}^q$ terms in the above. The choice of this form of position embedding is to make our formulation equivalent to how it is encoded in contemporary works \cite{Wang_2020_ECCV, ramachandran2019standalone} under the assumptions of Section~\ref{sec:relate_std_attention}. There may be other ways to encode position embeddings within our framework such as directly influencing the prior based on some distance measure $d(i, j)$ between the locations of units $i$ and $j$, as given by
\begin{align}
    \pi_{i,j} \propto \exp{(-d(i, j))}.
\end{align}

\section{Models, training and evaluation} \label{appx:models_training_evaluation}
Details of the interactive segmentation models used in the experiments and their training and evaluation procedures are provided below.

\subsection{Interactive segmentation model}
We use a single model to both predict an initial mask and correct it subsequently given an input image and annotator corrections. It takes a 3 channel input RGB image and 3 additional channels, one each to encode the object bounding box, positive and negative annotator corrections respectively. The object bounding box is specified using 2 clicks to roughly correspond to the box corners (top-left+bottom-right or bottom-left+top-right). These along with the positive and negative corrective clicks provided by the annotator are encoded as binary disks of radius 8 pixels following the findings in previous works \cite{Benenson_2019_CVPR, Maninis_2018_CVPR, Majumder_2019_CVPR}. We experiment using different architectures: HRNetV2+OCR \cite{Yuan_2020_ECCV} and DeepLabV3+ \cite{Chen_2018_ECCV}, backbones: ResNet-50 and ResNet-101 \cite{He_2016_CVPR},  and training datasets: SBD \cite{Bharath_2011_ICCV} and LVIS \cite{Gupta_2019_CVPR}) for specific experiments.

\subsection{Training}
All of our models are trained following a curriculum 
over three tasks. The first task is to predict a mask given an input image and object bounding box but empty corrective channels. The second task is to predict a mask given the image, bounding box and the corrective channels populated with randomly sampled clicks on the object foreground (positive) and background (negative). The third task is the corrective task, which is similar to the second task but with corrective channels containing corrective clicks randomly sampled from the false positive and negative error regions of the model's prediction. For both the second and third tasks, we randomly sample 1-3 clicks and 0-3 clicks for the positive and negative channels respectively. All our models are trained using RAdam optimizer \cite{liu_2020_ICLR} on a polynomial decay learning rate schedule (power=0.9) with a base learning rate of $10^{-4}$ decaying to $10^{-5}$ in 70 epochs and constant thereafter for a total of 150 epochs. The first and second tasks of our curriculum use 20 epochs each in a sequence and the remaining epochs are used for the third task. We use a batch size of 16 and train our models over 4 NVidia Volta 2-GPU nodes with 32GB of GPU memory each using Distributed Data Parallel framework in PyTorch. 

\subsection{Evaluation}
The trained models are evaluated on the GrabCut \cite{Rother_2004_ATG} and Berkeley \cite{Martin_2001_ICCV} datasets. Specifically, we plot the improvement in mask accuracy, i.e. mean IOU relative to ground truth, as a function of the number of clicks \cite{Benenson_2019_CVPR}, starting with the initial 2 clicks to specify a bounding box. For all the experiments, we simulate annotators by sampling the next corrective click from the largest error areas (positive and negative) obtained by comparing the ground truth mask with the model's current prediction. 
Note that for both training and evaluation, we crop the object bounding boxes with a finite padding around them without resizing the input image, as network input. We also add noise to the bounding box corners which introduces some variance into the evaluation metrics. In order to account for the variances in the crops, we repeat simulations over multiple trials (5 or 10) and report both the mean and the standard error across trials. We would like to point out that a better performing model reaches a certain mIoU faster, i.e. with fewer clicks.

\subsection{Other hyperparameters}
Apart from the training hyperparameters described above, we set the value of the query/key Gaussian precisions $\alpha_j$ to a constant value equal to $\frac{1}{\sqrt{d}}$, where $d$ is the query/key embedding dimension. This is similar to standard scaled dot-product attention. We use non-zero precisions $\beta_j$ only to train models with value propagation (Eq.~\eqref{eqn:value_propagation}). In that case they are all set equal to 0.1. 

Also, key adaptation and value propagation iterations use additional hyperparameters, specifically the priors $\theta_{\xi}$ and $\theta_{\mu}$. We set $\theta_{\xi}$ to 0 (ML update) unless otherwise specified. In order to choose an optimum value for $\theta_{\mu}$, we do a grid search over a set of 4 values: [0.1, 1, 10, 100] and choose the one that results in the least average number of clicks to reach a certain IOU (90\%) over all the images of a held-out set. It is possible that tuning these parameters specifically for each image might yield better results but we did not do this for simplicity and leave it as an exploration for future work.       

\section{Additional results using key adaptation}\label{appx:key_adaptation_results}
\begin{figure*}[htbp]
    \centering
    \subfloat[GrabCut]
    {
        \includegraphics[width=0.5\linewidth]{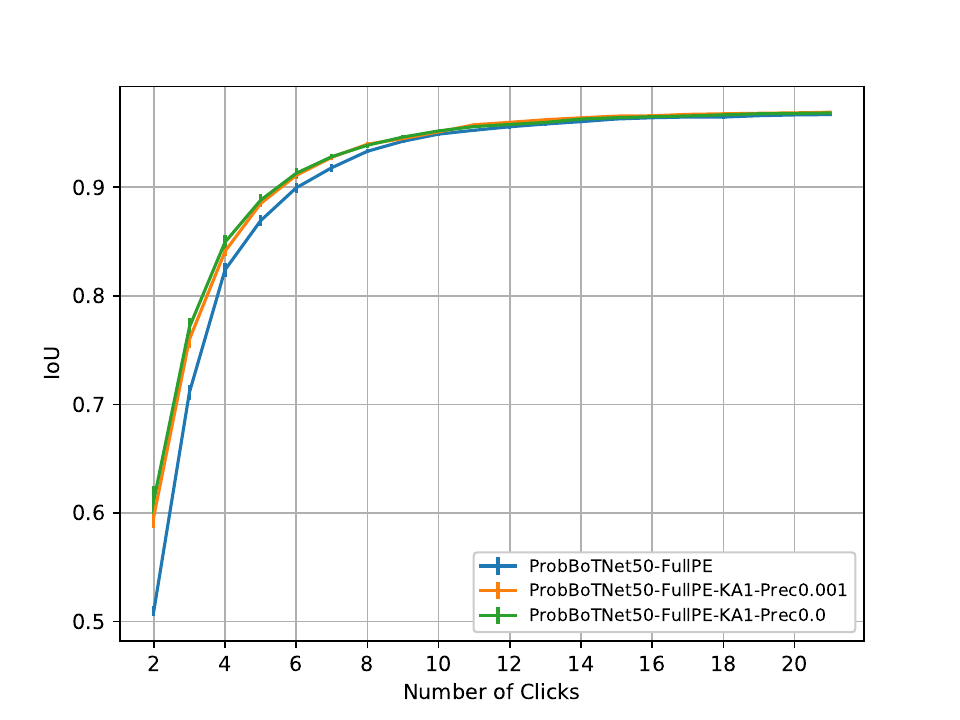}
        \label{fig:probatt_botnetbb_fullPE_KA_grabcut}
    }
    \subfloat[Berkeley]
    {
        \includegraphics[width=0.5\linewidth]{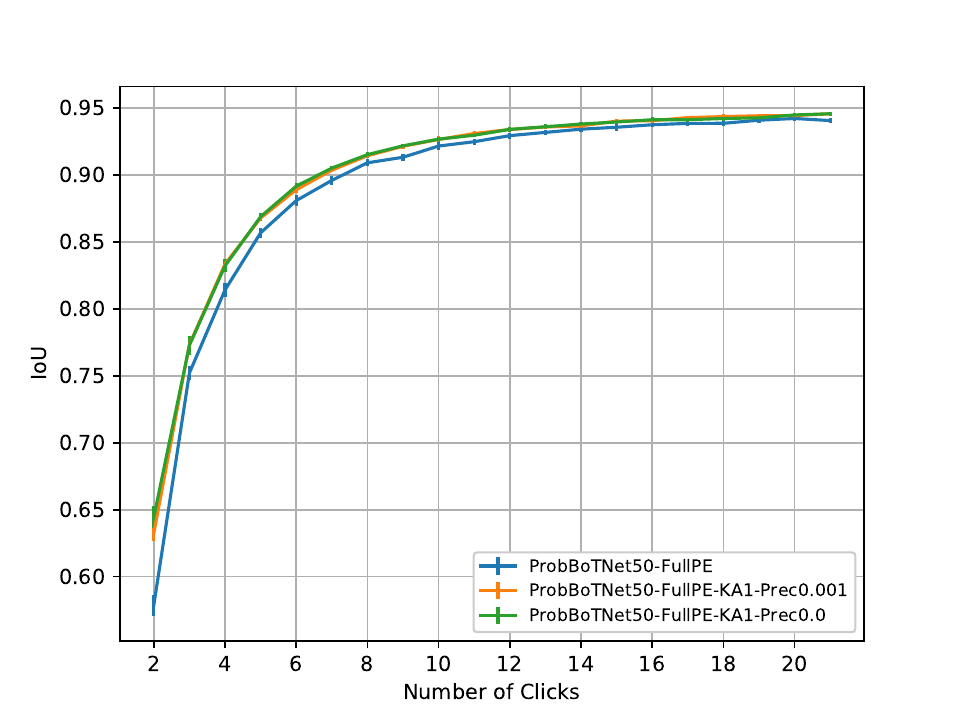}
        \label{fig:probatt_botnetbb_fullPE_KA_berkeley}
    }
    \caption{\textbf{Unsupervised key adaptation}. Mean IoU vs \#clicks with and without key adaptation (KA) on the GrabCut and Berkeley datasets. Probabilistic BoTNets with 
    full position encodings are evaluated without using KA or using 1 iteration of KA with two different prior precision (Prec.) values of 0.001 or 0.}
    \label{appx_fig:probatt_botnetbb_KA}
\end{figure*}

We provide additional results here (Fig.~\ref{appx_fig:probatt_botnetbb_KA}) using key adaptation within the probabilistic attention BoTNet50 architecture employing full position embedding: ProbBoTNet50-FullPE (Section~\ref{sec:probabilistic_attention_backbone}). See Section~\ref{sec:key_adaptation} for a discussion of the results.

\section{Additional results using value propagation}\label{appx:value_propagation_results}
\begin{figure*}[htbp]
	\centering
	\subfloat[GrabCut]
	{
			\includegraphics[width=0.5\linewidth]{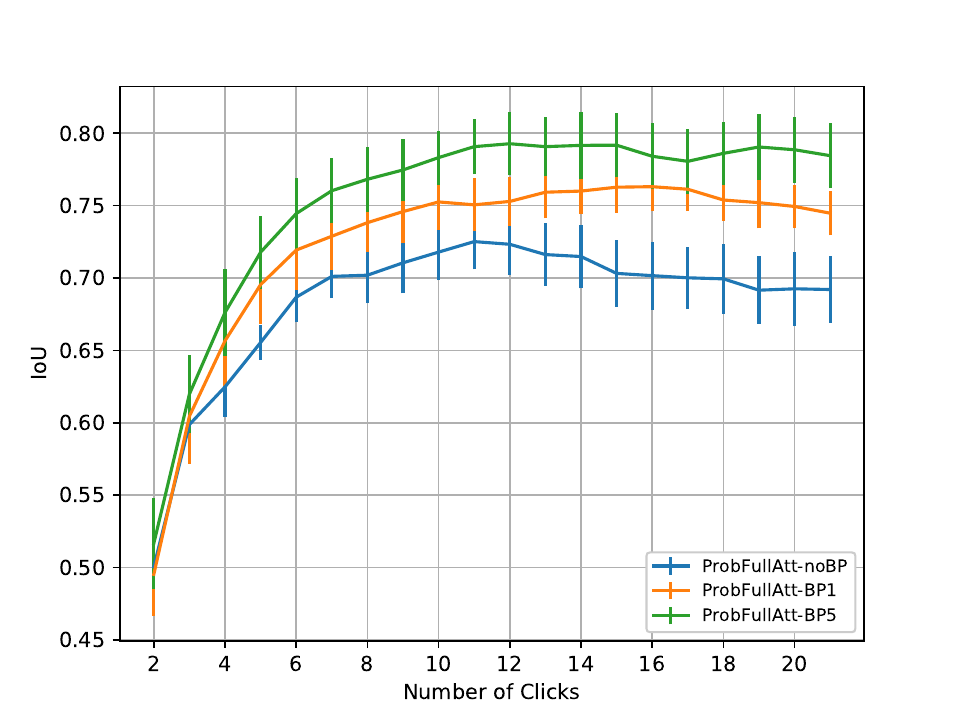}
			\label{fig:probfullatt_BP_grabcut}
	}
	\subfloat[Berkeley]
	{
			\includegraphics[width=0.5\linewidth]{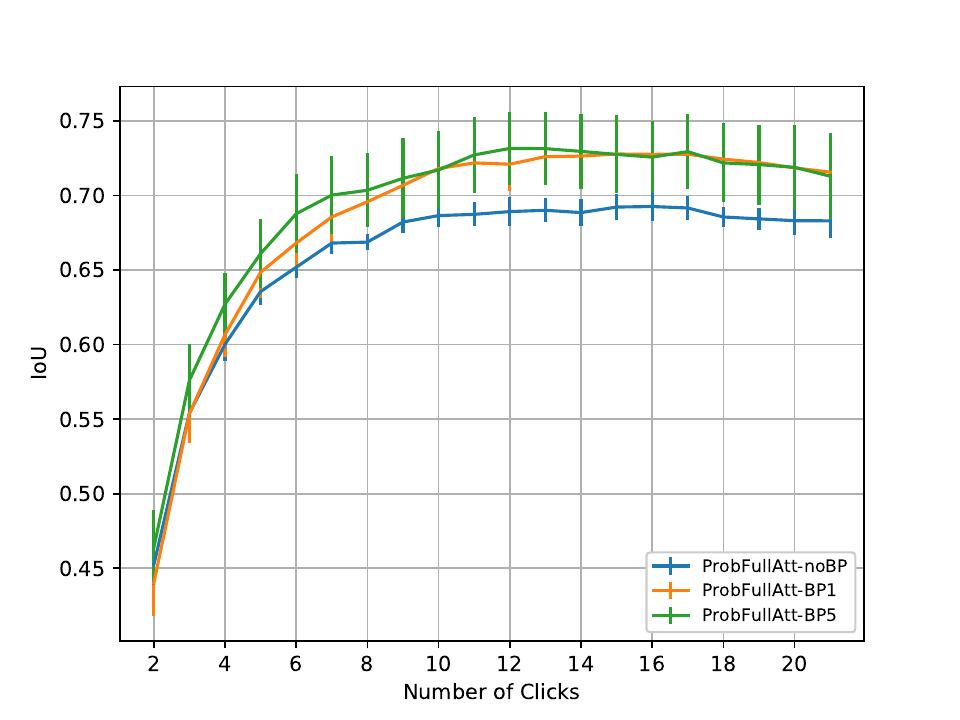}
			\label{fig:probfullatt_BP_berkeley}
	}
	\caption{\textbf{Effect of value propagation using full attention}. Mean IoU vs \#clicks with and without value propagation. We use 1 (BP1) and 5 (BP5) iterations of value propagation at the output self attention based classification layer of a network and test on the GrabCut (left) and Berkeley (right) datasets. }
    \label{fig:probfullatt_BP}
\end{figure*}

We conduct a small scale experiment to demonstrate the effect of value propagation using full attention instead of axial attention. For this experiment, we use a full self attention layer at the output of the network in place of the 1x1 conv classifier and feed in images at a resolution of 64 pixels. The corrective clicks are appended as additional inputs to this layer as described in Section~\ref{sec:probabilistic_attention_classifier}. The small input resolution allows us to work within the memory limits of current GPUs while being able to use full attention at the output layer \cite{Srinivas_2021_arxiv}. We use the Imagenet classification pre-trained HRNetV2+OCR \cite{Yuan_2020_ECCV} architecture and fine-tune it on SBD dataset \cite{Bharath_2011_ICCV} on the interactive segmentation task. Following the same protocol as in Section~\ref{sec:value_propagation_results}, the results are shown in Fig.~\ref{fig:probfullatt_BP}.

\end{document}